\documentclass[letterpaper, 10 pt, conference]{ieeeconf}  

\IEEEoverridecommandlockouts                              

\usepackage{graphicx}
\usepackage{multicol}
\usepackage{booktabs}
\usepackage{multirow}
\usepackage{tabularx}
\newcolumntype{Y}{>{\centering\arraybackslash}X}
\newcolumntype{L}[1]{>{\raggedright\let\newline\\\arraybackslash\hspace{0pt}}m{#1}}
\newcolumntype{C}[1]{>{\centering\let\newline\\\arraybackslash\hspace{0pt}}m{#1}}
\newcolumntype{R}[1]{>{\raggedleft\let\newline\\\arraybackslash\hspace{0pt}}m{#1}}

\usepackage{dingbat}

\usepackage{pifont}
\newcommand{\cmark}{\ding{51}}%
\newcommand{\xmark}{\ding{55}}%

\usepackage{hyperref}
\hypersetup{
    colorlinks=true,
    linkcolor=blue,
    filecolor=magenta,      
    urlcolor=blue,
    pdftitle={Overleaf Example},
    pdfpagemode=FullScreen,
    }
\usepackage{xcolor}

\usepackage{xspace}
\makeatletter
\DeclareRobustCommand\onedot{\futurelet\@let@token\@onedot}
\def\@onedot{\ifx\@let@token.\else.\null\fi\xspace}
\def\eg{\emph{e.g}\onedot} 
\def\ie{\emph{i.e}\onedot} 
 
\def\etc{\emph{etc}\onedot}

\makeatother

\definecolor{greencolor}{rgb}{0,0.8,0}

\definecolor{mycolour}{HTML}{000000}

\newcommand{\nsap}[0]{{MuBlE}\xspace} 
\newcommand{\nsemr}[0]{{CLIER}\xspace} 
\newcommand{\dataset}[0]{{SHOP-VRB2}\xspace} 
\newcommand{\ycbdataset}[0]{{YCB-VRB}\xspace} 

\title{\LARGE \bf
Closed Loop Interactive Embodied Reasoning for Robot Manipulation
}

\author{Michal Nazarczuk$^{1}$, Jan Kristof Behrens$^{2}$, Karla Stepanova$^{2}$, Matej Hoffmann$^{2}$ and Krystian Mikolajczyk$^{3}$
\thanks{$^{1}$Michal Nazarczuk is with Huawei Noah's Ark London
        {\tt\small michal.nazarczuk1@huawei.com}}%
\thanks{$^{2}$Jan Kristof Behrens, Karla Stepanova and Matej Hoffmann are with the Czech Technical University in Prague
        {\tt\small [jan.kristof.behrens, karla.stepanova]@cvut.cz}, }
\thanks{$^{3}$Krystian Mikolajczyk is with Imperial College London
        {\tt\small k.mikolajczyk@imperial.ac.uk}}%
\thanks{J.K.B., K.S., and M.H. were supported by the European Union under the project ROBOPROX (reg. no. CZ.02.01.01/00/22\_008/0004590). This work originated in the project Interactive Perception-Action-Learning for Modelling Objects (IPALM) (H2020-FET-ERA-NET Cofund-CHIST-ERA III / TAČR EPSILON, No. TH05020001)}
}

\begin{document}

\maketitle
\thispagestyle{empty}
\pagestyle{empty}

\begin{abstract}
Embodied reasoning systems integrate robotic hardware and cognitive processes to perform complex tasks, typically in response to a natural language query about a specific physical environment. This usually involves changing the belief about the scene or physically interacting and changing the scene (\eg sort the objects from lightest to heaviest). 
In order to facilitate the development of such systems we introduce a new modular Closed Loop Interactive Embodied Reasoning (CLIER) approach that takes into account the measurements of non-visual object properties, changes in the scene caused by external disturbances as well as uncertain outcomes of robotic actions. CLIER performs multi-modal reasoning and action planning and generates a sequence of primitive actions that can be executed by a robot manipulator. Our method operates in a closed loop, responding to changes in the environment. 

Our approach is developed with the use of MuBle simulation environment and tested in 10 interactive benchmark scenarios.
We extensively evaluate our reasoning approach in simulation and in real-world manipulation tasks with a success rate above 76\% and 64\%, respectively. 

\end{abstract}

\section{Introduction}
\label{sec:intro}
The research efforts in developing systems capable of high-level perception and reasoning have increased in recent years (\cite{Huang2019TransferableNavigation, Kamath2021MDETRUnderstanding, Kim2021ViLT:Supervision, Shridhar2020ALFREDTasks, garrett2020online, behrens2021embodied, goyal2024rvt, michal2024robotic, kruzliak2024interactive}), including models that utilize large pretrained vision-language models for action planning such as~\cite{huang2022ZeroShot, singh2023progprompt, kim24openvla}. However, the ability of these models to reason iteratively through complex problems is restricted by the necessity to ground the reasoning about robot policies in sensory observations and the robot state. Even if some of the above-mentioned reasoning agents are able to generate and execute a plan, either their execution, their planning, or both are open loop (\eg \cite{behrens2021embodied,michal2024robotic}). This means that they do not account for failures during the execution (open-loop execution) and cannot react to the knowledge acquired during execution, thus cannot solve long-term planning that requires to collect and maintain a set of sequential observations that change the action plan such as 'find the softest mattress' (open-loop planning). Furthermore, they typically reason only over visual observations and goals, which do not require physical measurements via interaction with the environment (see \cite{kruzliak2024interactive}, where exploratory actions to find physical object properties are selected using inference in a Bayesian network). Such methods are unable to robustly execute tasks such as passing the lightest bottle or finding the object with the coarsest surface. Although some recent models (e.g., \cite{goyal2024rvt}, \cite{yao2022react}, \cite{garrett2020online}) incorporate closed-loop reasoning, they focus on short-term tasks, e.g., manipulating an object. Robust and reactive long-horizon embodied reasoning remains an open problem. 

\begin{figure}
\centering
    \centering
     \includegraphics[width=0.83\linewidth]{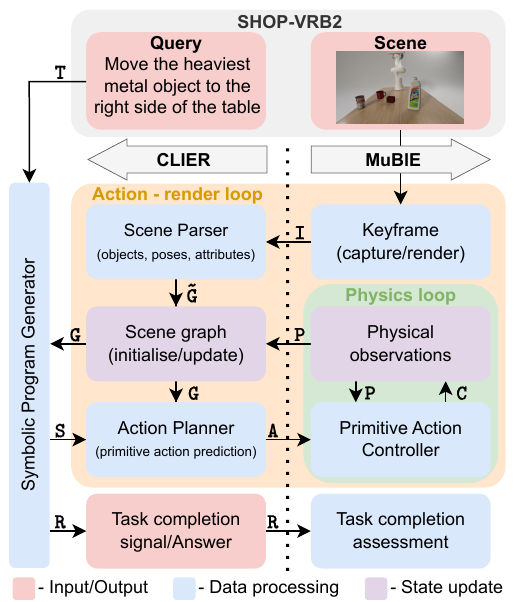}
    \caption{
        A diagram of CLIER reasoning implemented in MuBlE environment including interaction between the two and use of SHOP-VRB2 benchmark. Transferred data: $\mathtt{T}$ -- text of the query, $\mathtt{\tilde{G}}$ -- prediction of scene graph elements, $\mathtt{G}$ -- current scene graph, $\mathtt{S}$ -- subgoal (symbolic program requiring physical measurements), $\mathtt{I}$ -- image, $\mathtt{P}$ -- physical observations, $\mathtt{C}$ -- control signal, $\mathtt{A}$ -- primitive action to take, $\mathtt{R}$ -- returned result.
    }
    \label{fig:env_diag}
    \vspace{-1em}
\end{figure}

The main contribution of this article is the proposed Closed-Loop {neuro-symbolic} Interactive Reasoning framework~(\nsemr{})\footnote{The code and its documentation are released on the project webpage \href{https://michaal94.github.io/CLIER}{https://michaal94.github.io/CLIER}.} that can plan, evaluate, and execute a given task (\eg, \emph{Move the heaviest metal object to the right}) (see Fig.~\ref{fig:env_diag}). The highlight of our neuro-symbolic approach is that it can plan for perception of physical object properties (\eg, weight or stiffness) and account for new measurements to adjust the plan. CLIER processes visual inputs to identify objects and their poses, updating a scene graph before passing the information to an action planner. The planner selects the next robotic action, \eg, \textit{move} or \textit{weigh} (see Sec.\ref{s:physics}). Each action runs in a fast feedback loop (see \textit{Primitive Action Controller} in Fig.\ref{fig:env_diag}). We refer to all actions as primitive actions and any resulting observations—potentially from physical interactions—as measurements. Our reasoning agent is able to recover from a wide range of manipulation failures.

\begin{figure}[t]
    \centering
    \includegraphics[width=\linewidth]{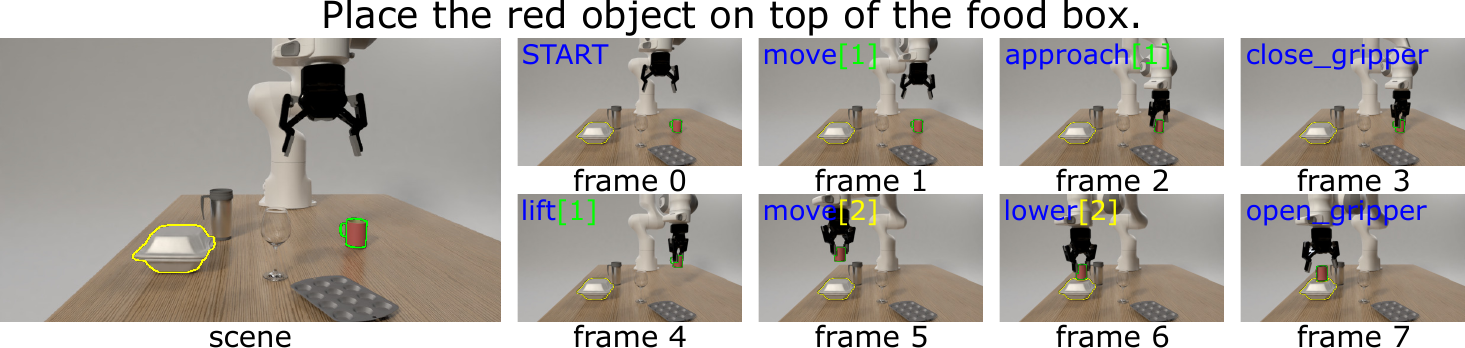}
    \caption{An example long-horizon manipulation task from \dataset{} implemented within the \nsap{} environment~\cite{muble}. A synthetic scene is rendered in Blender every keyframe, followed by the execution of the symbolic manipulation action planned by \nsemr{}.}
    \label{fig:teaser}
    \vspace{-1em}
\end{figure}

We evaluate our proposed Closed Loop Interactive Embodied Reasoning (\nsemr{}) within the \nsap{} simulation  environment~\cite{muble} using the \dataset{} synthetic benchmark (see an example task in Fig.~\ref{fig:sample_seq}) and validate it via comparative experiments between simulated and real-world \ycbdataset{} benchmark. Both of these datasets require combined reasoning between visual observations (recognising attributes of objects and their relations) and physical measurements (acquiring properties obtainable through interaction only, such as weight or stiffness).
\section{Related work}
\label{sec:relwork}

\noindent{\bf Symbolic approaches} to robot planning date back to the STRIPS planner \cite{fikes1972learning}, but there has been renewed interest in the possibility of incorporating unprocessed sensory, in particular visual, information. Several deep learning systems incorporate neuro-symbolic programs. CLEVR-IEP \cite{Johnson2017InferringReasoning} suggested predicting symbolic programs to be executed on a disentangled scene representation, followed by NS-VQA \cite{Yi2018Neural-SymbolicUnderstanding} and NS-CL \cite{Mao2019TheSupervision}. V2A \cite{Nazarczuk2020V2A-VisionLanguage} introduced a symbolic approach for robotic tasks using only the initial state of the scene and was evaluated in a real robotic environment~\cite{behrens2021embodied}, however, only open-loop single-step tasks were demonstrated. Symbolic approaches were also considered in Task and Motion Planning systems (TAMP) that were often based on the formal language PDDL \cite{Fox2003PDDL2.1Domains}. Classic TAMP approaches rely on predefined rules and known dynamic models \cite{Dantam2016IncrementalApproach, PackKaelbling2011HierarchicalNow, Migimatsu2019Object-CentricEnvironments}, but learning-based methods are increasingly replacing handcrafted heuristics in TAMP methods \cite{Chitnis2016GuidedHeuristics, Kim2019LearningRepresentation, Wang2018ActivePlanning}. PDDLStream \cite{Garrett_Lozano-Pérez_Kaelbling_2020} extends PDDL to add any sub-symbolic model in a black-box way, which also opens it to the use of learning-based methods. In~\cite{garrett2020online}, it was used for online replanning of short-horizon robotic tasks in the real world. Deep Visual Reasoning \cite{Driess2020DeepImage} directly predicts task plans by considering initial observation only. Regression Planning Networks (RPN) \cite{Xu2019RegressionNetworks} propose a task planning model performing regression in symbolic space using neural networks. RPN are expanded in \cite{Zhu2021HierarchicalGraphs} by utilising regression planning on scene graph representations to estimate the next subgoal based on the current scene and the scene graph, \ie solving only a subpart of the overall task that our method deals with. The interaction with the real world that follows a linguistic query (\eg {\it measure the weight of the mug}; {\it find the sofa}) is explored in {\bf embodied question answering approaches}~\cite{Das2018EmbodiedAnswering, behrens2021embodied, Yu2019Multi-TargetAnswering}. In \cite{behrens2021embodied}, language-conditioned visual reasoning is combined with manipulation. Navigation of agents based on vision and language was investigated in VLN \cite{Huang2019TransferableNavigation} or~\cite{liu2023grounding}. However, both approaches considered tasks that do not react to incoming measurements, \ie reasoning not conditioned on new observations. Recently, several works utilised large-language models (LLMs) to generate action plans for robot manipulation tasks (\eg~\cite{wu2023tidybot}, \cite{driess2023palm}, \cite{cheang2024gr}, \cite{kim24openvla}, \cite{black2024pi_0}). While they show impressive versatility and adaptability to novel environments, they suffer from a lack of fast-loop responsiveness, verifiability, and repeatability. LLMs require careful prompt engineering and filtering of outputs before the execution (\eg \cite{singh2023progprompt}). Furthermore, such approaches do not capture the task and trajectory details and do not enable fast-loop reactiveness. These problems have to be resolved before broader LLMs deployment. Using data created with \nsap{} environment, we train and evaluate a novel {\bf \nsemr reasoning agent} that can act on a natural language query in a real scene. Moreover, we present closed-loop reasoning experiments in simulation and real tabletop scenes. A highlight of our approach is that it plans for the perception of invisible object properties, \eg when stacking objects by weight. Moreover, unlike classical methods, our reasoning agent is executed on every keyframe, which allows it to recover from a wide range of manipulation failures (\eg, unsuccessful grasping or an object falling during manipulation).

\section{The Proposed CLIER Method} \label{sec:reasoning}

In this section, we present our closed-loop interactive reasoning approach (\nsemr{}) for long-horizon manipulation tasks that require embodied perception. The main modules are presented in Fig.~\ref{fig:env_diag} and an example task in Fig.~\ref{fig:pipeline}. \nsemr{} method is implemented within the \nsap{} simulation environment~\cite{muble}, which is introduced in Sec.~\ref{sec:muble}.

In the following we introduce the individual parts of our reasoning pipeline. \nsemr~starts by parsing the visual scene (see Sec.~\ref{s:parsing}) to create or update its semantic and geometric scene graph (see Sec.~\ref{s:graph}). As the reasoning agent we propose a multi-staged approach inspired by Neural Symbolic VQA approaches \cite{Nazarczuk2020SHOP-VRB:Perception, Yi2018Neural-SymbolicUnderstanding}. Simultaneously to the scene parsing, a symbolic program is extracted from a natural language instruction (see Sec.~\ref{s:symbolic_program}). Afterwards, the Symbolic Program Executor evaluates the symbolic program on the scene graph to select the subgoals that have to be achieved on the scene, \eg measuring the weight of the target when the graph node of the target's weight is empty (see Sec.~\ref{s:planner}). Such a subgoal is forwarded to the Action Planner, which takes into account the current state of the scene graph and the given subgoal, and selects a sequence of primitive actions to be executed in the environment. Each primitive action is executed within the physics loop (see Sec.~\ref{s:physics}). After each primitive action, the Action Planner reevaluates the next step within the action (render) loop (see Sect.~\ref{s:action}). The reasoning is completed when the Symbolic Program Executor finishes evaluation and either the answer is obtained, or the state of the scene matches the target state in the task completion modules.

\begin{figure*}
    \centering
    \begin{minipage}{0.5\linewidth}
    \includegraphics[height=1.9cm]{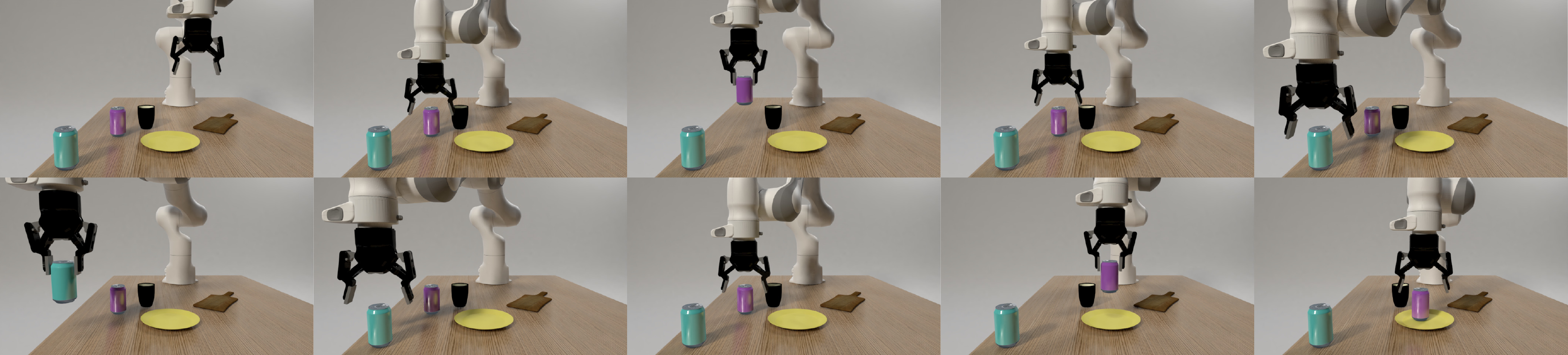}
    \end{minipage}
    \begin{minipage}{0.45\linewidth}
    \includegraphics[width=\linewidth]{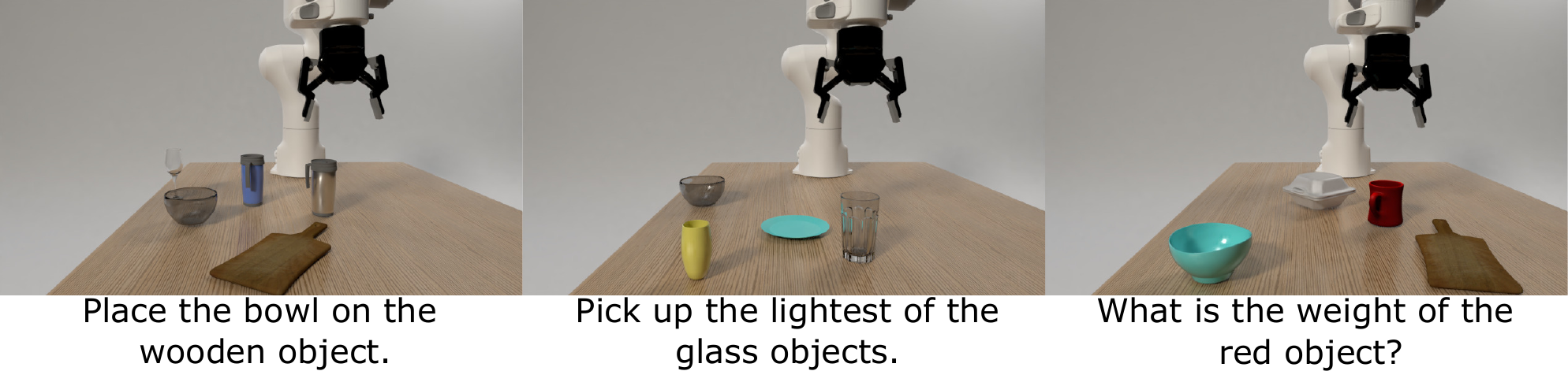}
    \end{minipage}
    \caption{(Left) Examples of simulated visual observations (selected frames) generated for the actions corresponding to the instruction: \textit{Stack the lightest of metal objects on the yellow object}. Note that metal cans were initially picked up to measure their weight. Further, according to the measurement taken, the heavier of the cans was put down and the lighter one was picked up and stacked on the yellow plate. (Right) Example simulated scenes and corresponding instructions in natural language generated with \nsap{} (in the dataset, instructions left to right belong to tasks 7, 3, and 1 in Tab.~\ref{tab:templates}).}
    \label{fig:sample_seq}
    \label{fig:scenes}
    \vspace{-1.5em}
\end{figure*}

\subsection{Scene Parser} \label{s:parsing}

A real scene can be captured by a camera, or a synthetic scene can be rendered by Blender based on a detailed scene description within the \nsap{} environment (see Sec.~\ref{sec:muble}). We capture/render the scene in every keyframe - \ie after a motion corresponding to the primitive action is finished (e.g. moving from object 1 to object 2, approaching grasping pose, closing the gripper, etc.). The keyframe-based approach allows us to use Blender for rendering frames, which have sufficient quality for sim2real transfer, and sufficient speed to allow for recovering from failures with the closed-loop approach. Example key frames rendered during execution of a task are presented in Fig.~\ref{fig:sample_seq}~(Left).
 
To parse the scene into a set of objects with attributes $\mathtt{\tilde{G}}$, we apply segmentation with Mask R-CNN~\cite{He2017MaskR-CNN}, followed by classification of object attributes with ResNet~\cite{He2016DeepRecognition} as in SHOP-VRB~\cite{Nazarczuk2020SHOP-VRB:Perception}
and pose regression with  CosyPose~\cite{Labbe2020CosyPose:Estimation}.

\subsection{Scene Graph} \label{s:graph}

A Scene graph $\mathtt{G}$ is generated from the parsed scene for each key frame $\mathtt{I}$ starting from the first timestep (see Fig.~\ref{fig:pipeline}). The Scene graph consists of feature vectors corresponding to each object in the scene, including categorical attributes for every object (\eg material, colour), along with their positions and orientations and physical properties (\eg weight, stiffness, elasticity, etc.). It also includes information on whether the object is in the gripper or whether the gripper is closed on it. Please note that the value of some of these properties might be empty (\eg unknown weight of the object).

We use geometrical heuristics to create or update the scene graph $\mathtt{G}$ for keyframe $\mathtt{I}$, similarly to \cite{Zhu2021HierarchicalGraphs}. These heuristics compare the relative positions of the objects and the end effector in order to define the following attributes: whether the object is in the gripper, whether the object is within the feasible grasp, whether the object is raised above the table, and whether the gripper is placed directly above the object.

\subsection{Symbolic Program Generator} \label{s:symbolic_program}

The first stage of our proposed reasoning approach inspired by neuro-symbolic VQA approaches \cite{Nazarczuk2020SHOP-VRB:Perception, Yi2018Neural-SymbolicUnderstanding} is a generation of a symbolic program. Symbolic Program Generator is implemented as a Seq2Seq network~\cite{Sutskever2014SequenceNetworks} translating the natural language query $\mathtt{T}$ into a sequence of symbolic programs similarly to~\cite{Nazarczuk2020SHOP-VRB:Perception}, trained with ground-truth. The generated program is in the CLEVR-IEP\,\cite{Johnson2017InferringReasoning} format.

Given the program sequence, it is evaluated on the current state of the scene graph $\mathtt{G}$, and either the task is finished, or a subgoal $\mathtt{S}$ necessary for task completion is produced, \eg measuring the weight of the target when the graph node of the target's weight is empty. When the Action Planner achieves the subgoal, the program is reevaluated and produces the next subgoal until task completion.

\subsection{Action planner} \label{s:planner}

The second stage of the reasoning is an Action Planner that for every keyframe $\mathtt{I}$ and a subgoal $\mathtt{S}$, given the scene graph $\mathtt{G}$, produces a sequence of primitive actions $\mathtt{[A]}$.

We employ a 2-layer transformer encoder as the action planner to predict the next primitive action $\mathtt{A}$ and its target, given the current scene graph $\mathtt{G}$ and the sub-goal $\mathtt{S}$, \eg measuring the weight of the target when the graph node of the target's weight is empty. We construct the input as the concatenation of the embeddings of the goal token and vectors describing all objects in the scene (including categorical properties, pose information, and relation to gripper based on the aforementioned heuristics). From this encoding, we predict the next action token, and select the target using the attention distribution over the objects training the network with ground-truth actions supervision. For example, when the symbolic program is queried for the task goal filter\_weight[heaviest], it encounters an empty field \textit{weight} in one of the scene graph nodes. This triggers the corresponding subgoal $\mathtt{S}$ for the Action Planner (\ie \textit{measure\_weight} with a pointer to the graph node of the object with the empty field).

Further, action planning regresses the task to a sequence of primitive actions for the current state of the scene. The Action Planner takes the current scene graph, the subgoal and its target as the input and produces the sequence of primitive actions that lead to executing the subgoal---\eg $\mathtt{move}$ -- $\mathtt{approach}$ -- $\mathtt{close\_ gripper}$ -- $\mathtt{lift}$ -- $\mathtt{weigh}$ for \textit{measure\_weight} subgoal). These actions are executed within the physics loop (see Sec.~\ref{s:physics}). Non-visual states, such as weight or stiffness are encoded in the scene graph after measurements, however, the transformer is provided only a binary flag indicating whether the property was measured or not and the reasoning is performed once all the measurements are completed. The binary flags simplify the process as the reasoning can be done simply over the binary values instead of continuous measurement outputs.

\subsection{Primitive actions and physics loop}
\label{s:physics}
To enable the planning of various manipulation tasks, we designed a set of primitive actions that can be easily extended in \nsap{}:
\begin{itemize}
    \item $\mathtt{move}$ -- moves the end effector towards a given target (\eg another object or part of the table),
    \item $\mathtt{approach}$ -- positions the end effector in a grasping position with respect to the target object
    \item $\mathtt{close\_ gripper}$, $\mathtt{open\_ gripper}$ -- to grasp or release,
    \item $\mathtt{lift}$, $\mathtt{lower}$ -- lifts or lowers the end effector,
    \item $\mathtt{weigh}$ -- weighs the object in the gripper,
    \item $\mathtt{squeeze}$ -- squeezes the object to measure its stiffness.
\end{itemize}

A specific controller is implemented to execute each primitive action $\mathtt{A}$ on the real robot or in the MuJoCo physics engine within the \nsap{} environment (see Sec.~\ref{s:env_physics}). The action of approaching to the grasp position entails a set of pre-coded grasp sequences. We utilize a simplified grasp simulation within \nsap{} -- see more details in~\cite{muble}. 

Each primitive action is executed within the physics loop. A control signal is generated at every step of the physics loop until the path is completed. It calculates the error for the current end effector pose and the planned trajectory and provides a control signal $\mathtt{C}$ to the environment. Every step of applying the control $\mathtt{C}$ affects the physics of the scene and generates a set of observations $\mathtt{P}$ (\eg pose of end effector, physical measurement, force in the gripper, \etc). Our method assumes that for each relevant physical property, there exists at least one primitive action that produces observations of this property (\eg $\mathtt{squeeze}$ for stiffness or $\mathtt{weigh}$ for mass).

\subsection{Action (render) loop} \label{s:action}

The action planner reevaluates the next step at every pass (every keyframe) of the action (render) loop that contains also scene graph generation (see Fig.~\ref{fig:env_diag}). For stability, we let the Action Planner always predict the full sequence of primitive actions until the subgoal completion. After the subgoal is completed (\eg, the weight field is populated in the target node of the graph), the Symbolic Program Executor resumes reasoning. It can either lead to producing the next subgoal (\eg, measuring the weight of another object if the query asks for comparison) or task completion.

The Primitive Action Controller implements a control loop that can execute action $\mathtt{A}$ on target object in MuJoCo physics engine or real robot. The physics loop, discussed in  Sec.~\ref{s:physics}), calculates physics and collects observations $\mathtt{P}$ (\eg pose of the end effector, physical measurement, force in the gripper, \etc). A control signal is generated in every step of the physics loop, until the path is completed. After the execution, a new key frame is captured in the real-world setup or rendered in simulation.

The reasoning is completed when either the answer is obtained from the program, or the state of the scene matches the target in task completion modules (\eg Weight of the object is returned or scene graph indicates that objects that were instructed to be stacked are on top of each other).

\section{Experimental setup}
\subsection{\nsap{} environment} \label{sec:muble}

We implement and evaluate our \nsemr{} method within the \nsap{} simulations environment for manipulation tasks~\cite{muble} (see Fig.~\ref{fig:env_diag}). It is built on $\mathtt{robosuite}$\,\cite{zhu2020robosuite}, a framework suitable for creating robotic environments in MuJoCo physics engine. The environment is equipped with high-quality rendering powered by Blender. It is designed for a tabletop scenario with a single robot manipulator and a gripper. It uses the same robots and grippers as $\mathtt{robosuite}$. It provides a set of object models in MuJoCo and their counterparts in Blender, along with templates to create new items.

\noindent{\bf{Physics}} \label{s:env_physics}
calculation uses the MuJoCo engine. With a user defined timestep, the control signal $\mathtt{C}$ is applied to the manipulator, the corresponding forces are calculated and applied to all objects in the scene. The simulator provides observations including visual and physical properties of objects and end effector, including measured values for non-visual properties such as weight, stiffness, or elasticity. Any custom sensor can be added to \nsap as in $\mathtt{robosuite}$.

\noindent{\bf{Data generation tools.}} \label{s:generation}
The \nsap{} environment also provides data generation tools: 1) \textbf{Scene Generator} places selected objects on the scene and sets their properties. It provides the rendered image of the scene (see Fig.~\ref{fig:scenes}~(Right)) along with full ground truth data, including segmentation masks as well as a depth map. 2) \textbf{Instruction Generator} given the scene, generates natural language instructions or tasks that require reasoning and interaction. The instruction generator also provides the symbolic program ground truth.

\begin{figure}
    \centering
    \includegraphics[width=1\linewidth]{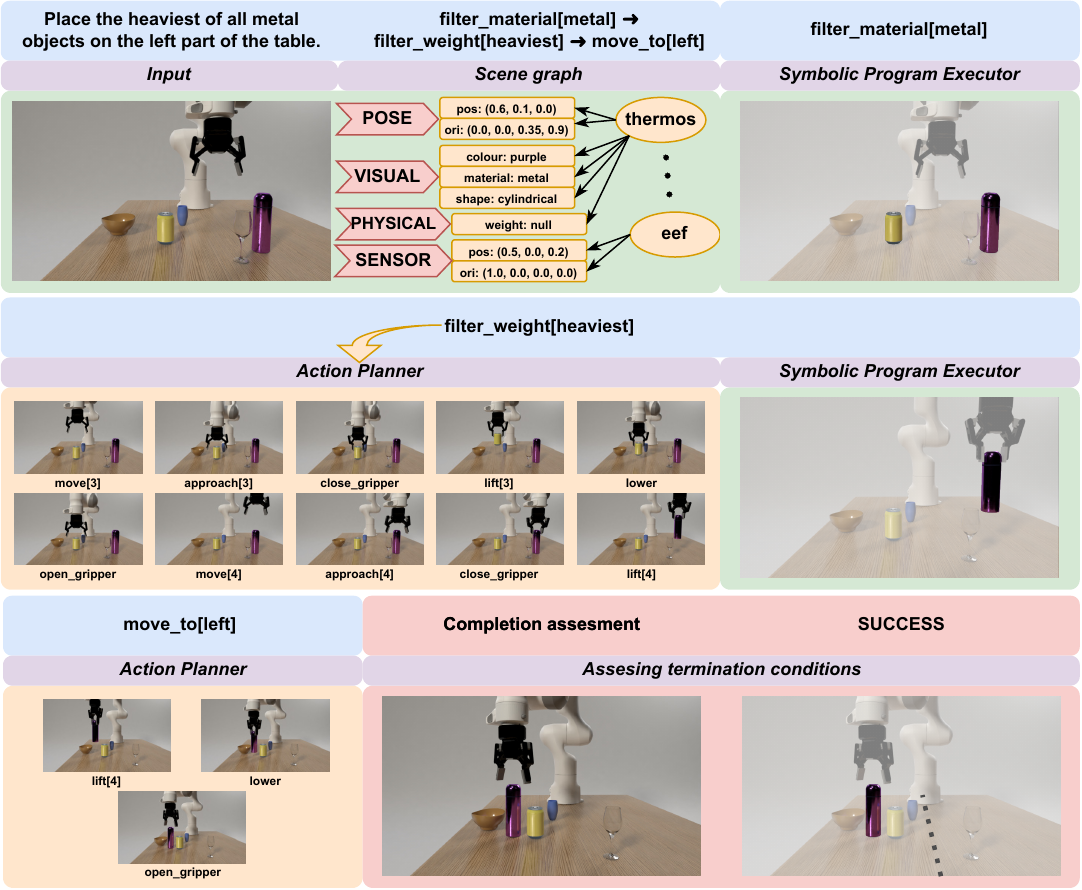}
    \caption{The \nsemr{} pipeline demonstrated on an example task from the \dataset{} dataset~\cite{muble}. Blue colour denotes a query and extracted symbolic programs, the output targets from programs are shown in the green, yellow denotes executing the sequence of primitive actions after having received a a subgoal from program (blue). Note that actions in yellow result in updates of the scene graph depicted in the top middle.}
    \label{fig:pipeline}
\vspace{-1em}
\end{figure}

\subsection{\dataset{} and \ycbdataset{} Datasets} \label{sec:dataset}

We evaluate the proposed \nsemr{} method on \dataset{} and \ycbdataset{} benchmarking datasets~\cite{muble} created in \nsap{}~\cite{muble} for training and benchmarking on long-horizon manipulation tasks. The datasets include a set of scenes with instructions to perform various tasks (\eg~\emph{Stack metal objects from heaviest to lightest}). Tasks are designed to enforce reasoning simultaneously on the visual observations (recognising attributes of objects and their relations) and continuous physical measurements (\eg measuring invisible properties), taking the feedback loop into account. Every example is accompanied with a ground truth sequence of actions in CLEVR-IEP \cite{Johnson2017InferringReasoning} format, along with visual observations and detailed scene graphs, including physical properties (see examples in Fig.~\ref{fig:sample_seq} and \ref{fig:pipeline}).

\subsubsection{Benchmarking tasks} \label{sec:instructions} 
There are 10 classes of benchmarking tasks (see Tab.~\ref{tab:templates}; more details can be found at~\cite{muble}). $\mathtt{OBJx}$ refers to a description of an object including visual properties, \eg \textit{the red object} presented in Fig.~\ref{fig:scenes}~(Right). $\mathtt{TPx}$ specifies left/right part of the table and $\mathtt{WSx}$ is a weight specifier (lightest/heaviest). Instructions have $5$ to $16$ words. The resulting sequences contain between $5$ (measure the weight of a single object) and $46$ (stacking several items according to weight) primitive actions.

\subsubsection{\dataset dataset} \label{s:benchmark}
\dataset{} dataset~\cite{muble} extends SHOP-VRB~\cite{Nazarczuk2020SHOP-VRB:Perception} dataset by introducing non-visible object properties and instructions that require closed-loop reasoning. {\bf \dataset scenes} include $12000$ realistically rendered scenes with typical household objects (similarly to \cite{Nazarczuk2020SHOP-VRB:Perception}). The scenes are generated as described in \nsap{}~\cite{muble}. Scene generator with $4$ to $5$ objects per scene. {\bf \dataset instructions} correspond to the tasks described in Sec.~\ref{s:benchmark} and are generated using \nsap{} Instruction Generator tool (see Sec.~\ref{sec:muble}). We assign one instruction to each scene. The instructions are randomly generated, however, they keep an equal distribution among the tasks from Table~\ref{tab:templates} and preserve training/test split as suggested in~\cite{muble}.

\begin{table}
    \caption{\scriptsize Instruction templates and benchmarking tasks in \dataset{}.}
    \begin{tabularx}{\linewidth}{rl}
    \toprule
    No. & Instruction Templates  \\
    \midrule
        1. & Measure the weight of the $\mathtt{OBJ1}$. \\
        2. & What is the weight of all $\mathtt{OBJ1}$s? \\
        3. & Pick up the $\mathtt{WS1}$ of all $\mathtt{OBJ1}$s. \\
        4. & Place the $\mathtt{OBJ1}$ on the $\mathtt{TP1}$ part of the table.\\
        5. & Remove all $\mathtt{OBJ1}$s from the $\mathtt{TP1}$ part of the table. \\
        6. & Place the $\mathtt{WS1}$ of all $\mathtt{OBJ1}$s on the $\mathtt{TP1}$ part of the table. \\
        7. & Stack the $\mathtt{OBJ1}$ on top of the $\mathtt{OBJ2}$. \\
        8. & Place the $\mathtt{WS1}$ of all $\mathtt{OBJ1}$s on top of the $\mathtt{OBJ2}$. \\
        9. & Stack the $\mathtt{OBJ1}$ on top of the $\mathtt{OBJ2}$ on top of the $\mathtt{OBJ3}$. \\
        10. & Stack all $\mathtt{OBJ1}$s from heaviest to lightest. \\
    \bottomrule
    \end{tabularx}
    \label{tab:templates}
\vspace{-1em}
\end{table}

\subsubsection{\ycbdataset{} dataset} \label{sec:benchmark_ycb_real}
{\bf \ycbdataset{} scenes} include 30 simulated benchmarking scenes with 4-5 objects randomly selected from the set of 9 YCB-Video \cite{Xiang2018PoseCNN:Scenes} objects sharing various visual/physical attributes. {\bf \ycbdataset{} Instructions} are generated similarly to \dataset{}. We generate 3 instructions per every template from Table~\ref{tab:templates}.

\subsubsection{Non-visual properties} Although the \dataset and \ycbdataset datasets contain only weight measurement as a representative example of estimating non-visual object properties through manipulation due to the ease of repeatability for other researchers, we also demonstrate performance of our method on tasks requiring stiffness measurements (see the accompanying video \cite{CLIER_youtube}) to show both the ease of extending the benchmark to other physical properties as well as the applicability of the proposed \nsemr{} method to various types of embodied actions. Including such a novel property only requires adding another instruction template for training and implementation of the control action to measure the given physical property.

\section{Results} \label{sec:results}
In this section we provide results for  \nsemr{}: 1) in  simulation for \dataset{} dataset (Sec.~\ref{sec:benchmark_shop}), and 2) on a set of 30 benchmarking scenes in simulation and in the real world with YCB objects~(Sec.~\ref{sec:benchmark_ycb}). 

\begin{table}
\begin{minipage}{\linewidth}
    \begin{center}
    \caption{\scriptsize (Left) Success rates for \nsemr{} method on \dataset{} (SHOP, sim) and YCB dataset (YCB sim/real).  (Right) Execution outcomes on \dataset{}, incl. successes (bold) and types of failures. See the webpage linked in Sec.~\ref{sec:intro} for more details and video illustration.}
    \label{tab:subtask}
    \end{center}
\end{minipage}
\centering
\begin{minipage}{\linewidth}
\begin{center}
\scriptsize
    \begin{tabularx}{0.9\textwidth}{l@{\hspace*{1mm}}c@{\hspace*{2mm}}c@{\hspace*{2mm}}c@{\hspace*{2mm}}c@{\hspace*{8mm}}l@{\hspace*{2mm}}c}
    \cmidrule[\heavyrulewidth](){1-4}\cmidrule[\heavyrulewidth](){6-7}
    Success [$\%$] & SHOP & \multicolumn{2}{c}{YCB\,\,\,} & & Failure type $[\%]$ & VRB \\
    Task type & Sim & Sim & Real & & Exit code & Sim \\
    \cmidrule(){1-4}\cmidrule(){6-7}
    Weight single & $74.0$ & $66.7$ & 88.9 & & \textbf{Correct answer} & $13.9$ \\
    Weight multi &  $65.0$ & $100$ & 66.7 & &\textbf{Task success} & $30.0$ \\
    Pick up weight & $49.0$  & $100$ & 88.9 & & Task failure & $0.1$ \\
    Move single &  $76.0$ & $66.7$ & 100 & & Execution err & $14.4$ \\
    Move multi & $47.0$  & $100$ & 44.4 & & Loop detected & $10.8$ \\
    Move weight & $23.0$  & $100$ & 100 & & Physics err & $3.5$ \\
    Stack & $56.0$  & $66.7$ & 66.7 & & Program err & $4.5$ \\
    Stack weight & $31.0$  & $33.3$ & 22.2 & & Recognition err & $9.6$ \\
    Stack three & $0.0$  & $66.7$ & 0.0 & & Output error & $0.6$ \\
    Order weight & $18.0$  & $66.7$ & 100& & Scene inconsistent & $12.6$ \\
    \cmidrule[\heavyrulewidth](){1-4}
    Overall & $43.9$ & $76.7$ & $64.4$ & &  & \\
    \cmidrule[\heavyrulewidth](){1-4}
    \end{tabularx}
\end{center}
\end{minipage}
\vspace{-2em}
\end{table}

\noindent{\bf Evaluation metrics} are the rate of successful task execution (\eg stacking) and percentage of correctly answered questions (\eg weight query). We also provide accuracy split into task types presented in Tab.~\ref{tab:templates}. Finally, for the inference tool of \nsemr{} we provide a classification of incorrect attempts, including errors of execution, incorrect scene recognition, loop detection, timeout, and inconsistency in object tracking. 

The latency of various modules is the following. CosyPose inference is ~0.8s, attributes recognition: ~0.15s, transformer action planner: ~0.02s (as run on 2080Ti). The prediction of the next action takes around 1s as measured on the hardware. These are reasonable delays given that these modules are deployed at keyframes.

\subsection{\dataset experiments} \label{sec:benchmark_shop}
Results for \nsemr{} are presented in Table \ref{tab:subtask} (Left) which shows success rates for various tasks. We observe high success rates for the tasks that include manipulation of single objects (weighing or moving one object) and a strong decrease in accuracy for multi-object manipulation (stacking more objects, with preceding manipulation for weighting). Further, we identify the most common reasons for failure in Table \ref{tab:subtask} (Right), which reports the distribution of error codes from the environment. Execution error ($14.4\%$) accounts for failures in execution of primitive actions which may arise from inaccurate scene description (\eg typically object pose). Scene inconsistency ($12.6\%$) refers to mistakes in tracking objects IDs between frames. Loop detection ($10.8\%$) arises when primitive action chains are repeated (\eg when approaching object to grasp with misaligned position). Note that this experiment uses simple, ResNet based pose estimation in contrast to more accurate CosyPose~\cite{Labbe2020CosyPose:Estimation} that was trained for YCB objects but not for \dataset{} objects. Finally, we believe that the overall accuracy of \textbf{$43.9\%$} on \dataset{} indicates that this data forms a challenging benchmark for evaluating visual and interactive reasoning.

\subsection{\ycbdataset{} and real-world experiments} \label{sec:benchmark_ycb}

In this section, we report comparative results for real YCB scenes that mirror 30 YCB scenes introduced in Sec.~\ref{s:benchmark}.  

\begin{figure}
    \centering
    \includegraphics[width=0.91\linewidth]{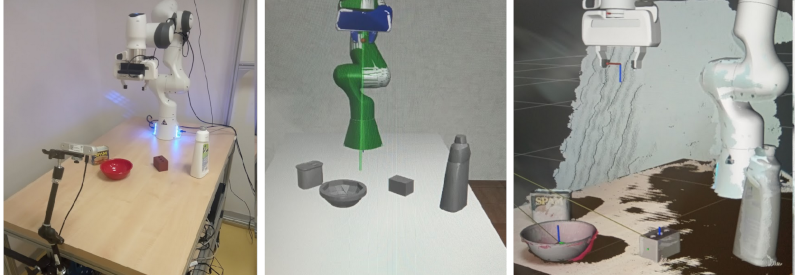}
    \caption{The real setup with YCB objects (left), corresponding MuJoCo simulation using estimated poses (middle), and RViz visualisation of colored pointcloud with overlaid gray models detected by CosyPose (right).}
    \label{fig:real_robot}
    \vspace{-1em}
\end{figure}

\noindent{\bf Experimental Setup} includes the Franka Emika Panda~\cite{franka} arm with $7$ degrees of freedom and a $2$-finger parallel gripper. An (extrinsically calibrated) Intel Realsense D455 camera is set to face the robot and capture the front view of the scene (see Fig.~\ref{fig:real_robot}). To allow \nsemr{} to control the real robot, we developed a thin ZMQ~\cite{zeromq} based communication layer that emulates the same interface as for the simulated robot (\ie, motion generation using position-based servoing). Special purpose skills include measuring stiffness via squeezing the object with different forces and measuring deformation or weight via lifting and joint torque differences. Note that the weighing has a relative error of less than $\pm 10\%$ in the range $0-200$~g compared to the ground truth weight. The stiffness measurement has a coefficient of variation ranging from $1.6-3.4$~\%, which is highly repeatable. However, we were not able to obtain ground truth data for stiffness.

\noindent{\bf Sim2Real.} Our method uses raw RGB rendered images to train the attribute recognition module (see Fig.~\ref{fig:env_diag}). This is not further tuned between the YCB sim and real experiments. The same reasoning model (see Sec.~\ref{sec:reasoning}) is evaluated in 30 synthetic and real scenes using the same  CosyPose~\cite{Labbe2020CosyPose:Estimation} on RGB images. The reasoning modules operate on the scene graph and thus are agnostic to visual input. 

\noindent{\bf Results.} We mirror experiments with real scenes in our simulation environment. To evaluate the contribution of different modules we perform testing along with ablations by replacing them with ground truth (GT) data, \ie 1) reasoning success with GT pose and  GT attributes 2) pose inference with GT attributes and GT reasoning, 3) pose inference and attribute recognition with GT reasoning, 4) full inference without using GT. The results for all these combinations are in Table \ref{tab:simreal}, where the success is the average across 3 runs of each experiment. Note that the Seq2seq network in the symbolic program generator is not included in the table as it performed with $100\%$ accuracy, therefore the error for case 1) comes from the transformer action planner. We observe a very similar performance between simulation and real world which validates the usefulness of high-quality vision and physics in \nsap{}. The performance for individual tasks with YCB scenes in simulation and the real world is reported in Tab.~\ref{tab:subtask}(Left). Sim and Real differences for YCB can be attributed mostly to CosyPose object detection and its robustness to occlusion.

\noindent{\bf Typical execution errors.}
Our closed loop reasoning approach is able to recover from various execution errors such as dropping or moving an object (see video \cite{CLIER_youtube}). Typical errors during the real world execution are caused by uncertainty in pose estimation. This leads to various cases:  1) grasp succeeds because object's real pose is withing the margin for error. 2) a different grasp is achieved \eg over a rim instead over diameter of an open cylindrical object. 3) the object slips out when closing the gripper. 4) the gripper collides with the object during the approach and causes a robot error (safety stop reflex). The robot can recover from all except case (4) or when the object is pushed out of the workspace in (3). A varied pose of the object in the gripper (i.e., (1) and (2)) occasionally influences subsequent tasks such as stacking or weighing. The detection of occluded objects is also prone to failures. In particular, occlusion from the gripper when grasping is an issue when updating the scene graph. We use the last estimated object pose and leverage the information of the gripper holding `something' to overcome these pose estimation errors. 

\noindent{\bf Joint limits and singularities.} Cartesian control of the robot end-effector without taking the joint configuration into account can lead to degenerate configurations. These include 1) joint limits, 2) borders of the workspace, 3) self-collisions, and 4) alignment of several joint axes. To avoid running into joint limits, we start each experiment in a good standard pose, and the robot always rotates to a neutral down-facing end-effector orientation before moving via the smallest rotation to the goal rotation. Self collisions are avoided by the Panda controller and are counted as \textit{robot errors}.  

\begin{table}
\begin{center}
\caption{\scriptsize Success rate for different setups (inference) in simulation and real world. \textbf{C} - CosyPose, \textbf{A} - attributes recognition, \textbf{R} - reasoning, \cmark - module used, \xmark - module substituted with ground truth information.}
\label{tab:simreal}
\scriptsize
\begin{tabular}{ccclcc}
\toprule
\textbf{C} & \textbf{A} & \textbf{R} &  & Simulation & Real \\
\midrule
\xmark & \xmark & \cmark & & $86.7\%$ & $86.7\%$ \\
\cmark & \xmark & \xmark & & $90.0\%$ & $80.0\%$ \\
\cmark & \cmark & \xmark & & $90.0\%$ & $76.7\%$ \\
\cmark & \cmark & \cmark & & $76.7\%$ & $64.4\%$ \\
\bottomrule
\end{tabular}
\end{center}
\vspace{-2em}
\end{table}

\section{Conclusions} \label{sec:conclusions}

In this paper, we presented our Closed-Loop Interactive Embodied Reasoning (\nsemr{}) for robotic manipulation tasks. \nsemr{} is able to incorporate observations of both visual and physical attributes of the manipulated objects into long-term reasoning. Capturing data from visual and physical measurements in the shared scene graph enables the symbolic reasoning approach and simplifies the implementation of the closed loop approach via keyframe-based reasoning. The \nsemr{} utilises  \nsap{} environment~\cite{muble} that incorporates MuJoCo physics simulation with high-quality renderer and enables generation of multi-modal demonstration data for robotic manipulation tasks. The results from simulated and real-world experiments showed its ability to successfully transfer between the simulated and real environment and to recover from various errors or changes in the scene.

\bibliographystyle{IEEEtran}
\bibliography{references}

\end{document}